\documentclass[runningheads]{llncs}

\usepackage{caption} 
\usepackage{graphicx}
\usepackage{multirow}
\usepackage[section]{placeins}
\usepackage[normalem]{ulem}
\useunder{\uline}{\ul}{}
	
\begin{document}
\title{Difficulty Classification of Mountainbike Downhill Trails utilizing Deep Neural Networks}
\titlerunning{Difficulty Classification of Mountainbike Downhill Trails}

\author{Stefan Langer, Robert M\"uller, Kyrill Schmid and Claudia Linnhoff-Popien}
\authorrunning{Stefan Langer et al.}

\institute{Mobile and Distributed Systems Group\\LMU Munich\\
\email{\{stefan.langer,robert.mueller,kyrill.schmid,linnhoff\}@ifi.lmu.de}\\
}
\maketitle

\begin{abstract}
The difficulty of mountainbike downhill trails is a subjective perception. 
However, sports-associations and mountainbike park operators attempt to group trails into different levels of difficulty with scales like the \textit{Singletrail-Skala} (S0-S5) or  colored scales (blue, red, black, ...) as proposed by \textit{The International Mountain Bicycling Association}.
Inconsistencies in difficulty grading occur due to the various scales, different people grading the trails, differences in topography, and more.
We propose an end-to-end deep learning approach to classify trails into three difficulties easy, medium, and hard by using sensor data.
With mbientlab Meta Motion r0.2 sensor units, we record accelerometer- and gyroscope data of one rider on multiple trail segments.
A 2D convolutional neural network is trained with a stacked and concatenated representation of the aforementioned data as its input.
We run experiments with five different sample- and five different kernel sizes and achieve a maximum Sparse Categorical Accuracy of 0.9097.
To the best of our knowledge, this is the first work targeting computational difficulty classification of mountainbike downhill trails.

\end{abstract}
\section{Introduction}
Mountainbiking is a popular sport amongst outdoor enthusiasts, comprising many different styles.
There are styles like cross country riding, characterized by long endurance rides, styles like downhill riding, characterized by short, intense rides down trails, and more ~\cite{ucidisciplines}.
Mountainbiking, as it is known today, originated in the US in the 1970s and since then went through various levels of popularity ~\cite{gaulrapp2001injuries}.
Official, competitive riding started in the 1980s with the foundation of the \textit{Union Cycliste Internationale (UCI)}, followed by the first World Championship in 1990 ~\cite{impellizzeri2007physiology}. 
In this work, we focus on the difficulty classification of mountainbike downhill trails and do not take into account uphill or flat sections of trails.
There are multiple approaches in trail difficulty classification, whereby a color-inspired grading is most commonly used ~\cite{imbarating,britishcyclinggrades,schymik2008singletrail}.
The \textit{International Mountain Bicycling Association (IMBA)} proposes a trail difficulty rating system comprised of five grades, ranging from a green circle (easiest) to a double black diamond (extremely difficult) ~\cite{imbarating}. 
In addition, the \textit{IMBA Canada} offers a guideline on how to apply those gradings to mountainbike trails ~\cite{imbaguidelines}.
\textit{British Cycling} also propose a colored difficulty scale, including four basic grades from green (easy) to black (severe) with an additional orange for bike park trails ~\cite{britishcyclinggrades}.
Inspired by  rock climbing difficulty grading, as well as ski resort gradings, Schymik et al. created the \textit{Singletrail-Skala}, containing three main difficulty classes (blue, red, black) and a more fine granular six grades ranging from S0 to S5 ~\cite{schymik2008singletrail}. 
Trails on \textit{Openstreetmap} ~\cite{openstreetmap} are rated with respect to the IMBA grading as well as the \textit{Singletrail-Skala}, wheareas the latter also describes tracks which are not specificly made for mountainbiking ~\cite{osmsingletrails}. 
Due to factors like the various scales, different people grading the trails or differences in topography, estimating the difficulty of mountainbike trails consistently is not an easy task.
This work aims to make mountainbike track difficulty assessment less subjective and more measurable.
In order to do so, we collect acceleration-, as well as gyroscope-data from multiple sensor units that are connected to the mountainbike frame as well as the rider.
Because we do not collect data in dedicated mountainbike parks, but on open trails (hiking paths among others), we decided to use the three main difficulties given by the \textit{Singletrail-Skala} as the set of labels.

\begin{table}[]
\centering
\setlength{\tabcolsep}{0.5em} 
{\renewcommand{\arraystretch}{1.2}
\resizebox{\textwidth}{!}{%
\begin{tabular}{|l|l|l|l|}
\hline
Colored grading & Fine grading & Label & Description                                                                                                                                                    \\ \hline
blue            & S0, S1       & 0     & \begin{tabular}[c]{@{}l@{}}Easy. \\ Mostly solid and non-slip surface. \\ Slight to moderate gradient. \\ No switchbacks. \\ Basic skills needed.\end{tabular}         \\ \hline
red             & S2           & 1     & \begin{tabular}[c]{@{}l@{}}Medium. \\ Loose surface, bigger roots and stones.\\ Moderate steps and drops.\\ Moderate switchbacks.\\ Advanced skills needed.\end{tabular} \\ \hline
black           & S3+          & 2     & \begin{tabular}[c]{@{}l@{}}Hard. \\ Loose surface, slippery, big roots and stones.\\ High drops.\\ Tight switchbacks.\\ Very good skills needed.\end{tabular}           \\ \hline
\end{tabular}%
}
}
\caption{Mapping of the \textit{Singletrail-Skala} grades to the labels used in our data set. Label 0 describes easy trails, label 1 medium trails, and label 2 hard trails.}
\label{table:coloredgrading}
\end{table}

Table~\ref{table:coloredgrading} gives an overview of the three grades blue, red and black.
Schymik et. al ~\cite{schymik2008singletrail} define the difficulties as follows:
Blue describes easy trails, comprising the grades S0 and S1.
Red describes medium trails and is equal to the grade S2.
Black describes all difficulties above and can be considered hard.
\textit{Openstreetmap} provides difficulty classifications for all trails on which this dataset is collected ~\cite{osmsingletrails}.
We then train a 2D convolutional neural network with a stacked and concatenated representation of the aforementioned data as its input.
Thereby we can grade sections of downhill trails regarding their difficulty.

\subsection{Related Work}
For training purposes, mountainbikes of professional athletes get set up with telemetry technology, such as BYB Telemetry's sensors ~\cite{bybtelemetry}.
Their sensors are connected to the suspension fork as well as the suspension shock and measure the movement of each.
Stendec Data extends those capabilities and adds sensors for measuring brake pressure and acceleration in order to capture braking points, wheel movements, and more ~\cite{stendecracing}.
However, the two systems mentioned above are expensive and hard to get.
Therefore, we use mbientlab Meta Motion sensor units to capture acceleration and gyroscope data.
Ebert et al. ~\cite{ebert2017automated} automatically recognized the difficulty of boulder routes with mbientlab sensor units.
To the best of our knowledge there is no scientific work regarding the difficulty classification of mountainbike trails using accelerometers or gyroscopes yet.
However, there has been a great amount of work done in the field of activity recognition with acceleration data ~\cite{ravi2005activity,bao2004activity,kwapisz2011activity,lara2012survey,yang2015deep,zeng2014convolutional,ronao2016human}.
Many of those approaches make use of classical machine learning methods ~\cite{ravi2005activity,bao2004activity,kwapisz2011activity,lara2012survey,preece2008comparison}.
S. J. Preece et al. ~\cite{preece2008comparison} compare feature extraction methods for activity recognition in accelerometer data.
Ling Bao et al. ~\cite{bao2004activity} classify activities using custom algorithms and five  biaxial acceleration sensors worn simultaneously on different parts of the body.
Furthermore, there has been a noticeable shift towards deep learning approaches in recent years ~\cite{yang2015deep,zeng2014convolutional,ronao2016human}.
Fernando Moya Rueda et al. ~\cite{moya2018convolutional} use multiple convolutional neural networks which they concatenate in a later stage with fully connected layers.
Zeng et al. utilize a 1D convolutional neural network, treating each axis of the accelerometer as one channel of the initial convolutional layer ~\cite{zeng2014convolutional}.
In a survey by Jindong Wang et al., the authors give an overview of state-of-the art deep learning methods in activity recognition ~\cite{wang2019deep}.
The authors claim that deep learning outperforms traditional machine learning methods and has been widely adopted for sensor-based activity recognition tasks.

\section{The dataset}

\subsection{Collecting and labeling data}

\begin{figure}
\includegraphics[width=\textwidth]{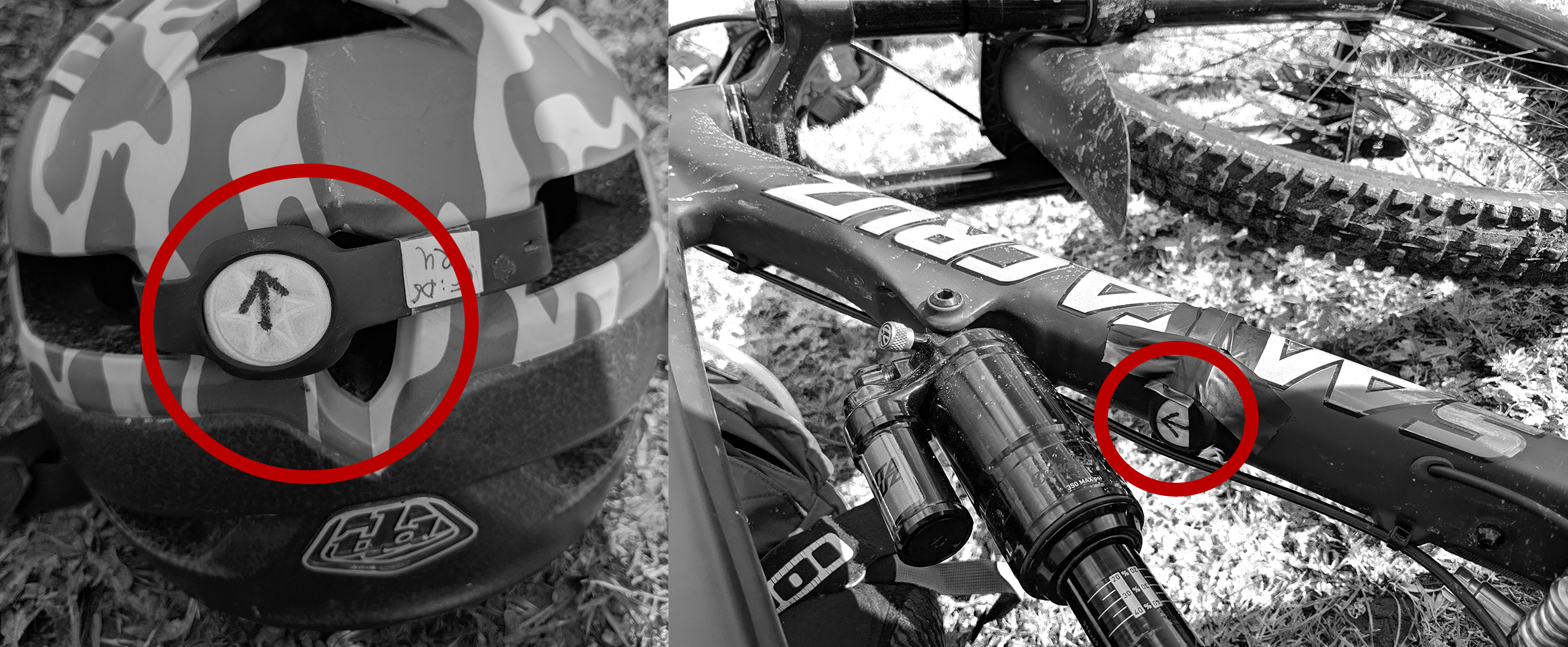}
\caption{Mounting point on the helmet (left), mounting point on the downtube of the frame (right)}
\label{fig1}	
\end{figure}

Instead of working with dedicated mountainbike telemetry systems, we use mbientlab Meta Motion r0.2 sensor units to record data ~\cite{mbientlab}.
Those units contain multiple sensors, including an accelerometer as well as a gyroscope.
Mbientlab sensors offer a Bluetooth Low Energy interface to which an Android or iOS application can be connected.
The rider is equipped with two sensor units.
Fig.~\ref{fig1} visualizes the mounting points of the mbientlab sensor units.
One unit is connected to the downtube of the mountainbike, the other one to the back of the rider's helmet.
For each recording, the sensors are facing the same direction to keep the axes layout consistent.
The accelerometer creates datapoints in three axes (x, y, z) in the unit g (equals $9.80665 m/s^2$) with a frequency of 12.50Hz.
The gyroscope creates datapoints in three axes (x, y, z) in the unit deg/s with a frequency of 25.00Hz.
We synchronize the starting points of the recordings and linearly interpolate missing datapoints to reach a constant frequency of 25.00Hz for all sensors.

Labeling of the data happens after the actual data collection process.
We record every downhill ride with an action camera (mounted to the rider's chest), synchronize the video with the data recordings, and manually label subsections of the trail.
For the majority of subsections on open trails, we use the difficulty grading provided by \textit{Openstreetmap}.
Those gradings are made visible in mountainbike specific \textit{Openstreetmap} variants and can also be found in the (XML-like) \textit{.OSM} exports of an area.
One "way" node (which describes a trail) then includes another node "tag", comprising the difficulty description.
For subsections that the \textit{Singletrail Skala} would consider to not represent this difficulty (as per their description), we up- or downgrade the difficulty label.
Downgrading mostly occurs for fireroads or other very easy sections, upgrading for particularly steep or tight sections.

\FloatBarrier

\subsection{Input data representation}

\begin{figure}
\includegraphics[width=\textwidth]{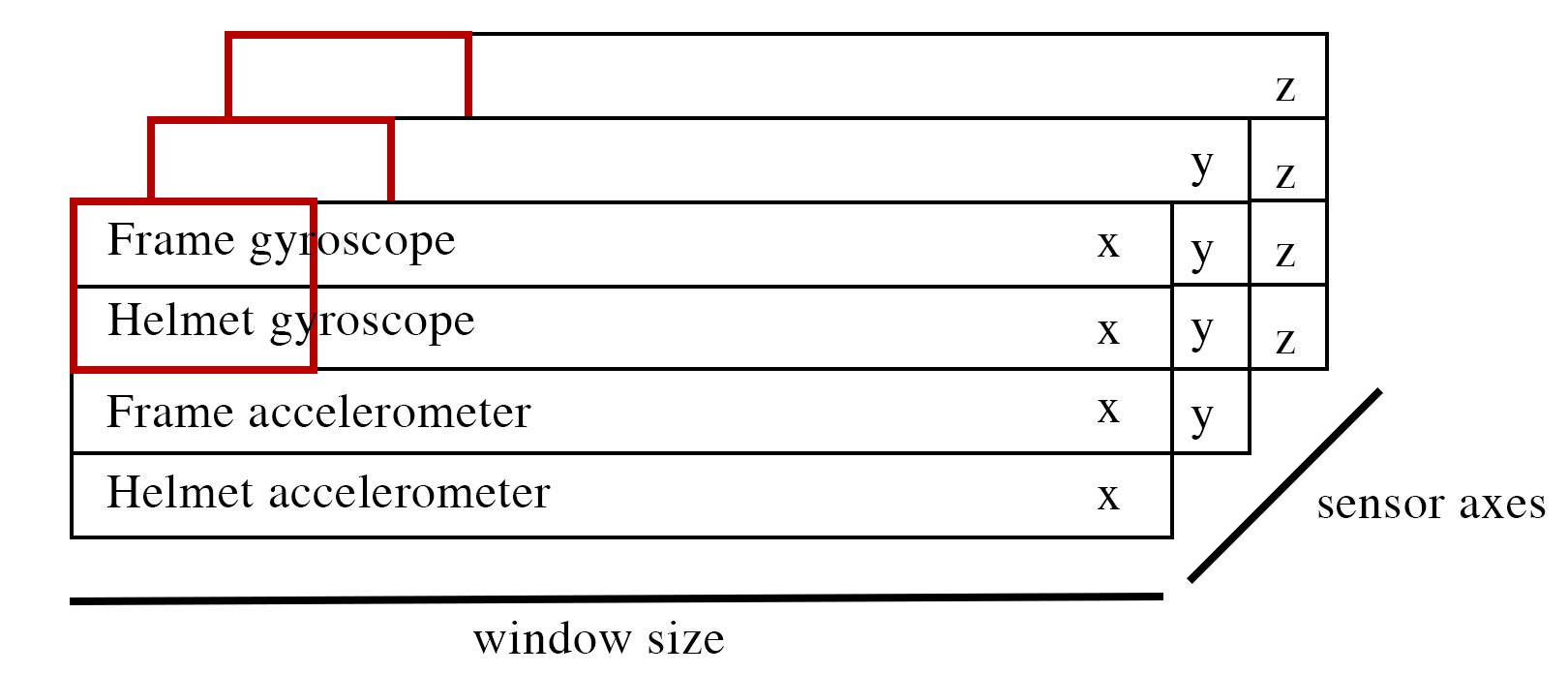}
\caption{Input shape of one sample comprising four sensors with three dimensions (axes) each}
\label{fig2}	
\end{figure}

For each ride, we collect data with two sensor units.
Every unit provides data for the accelerometer and the gyroscpope sensors.
Each sensor generates datapoints for three axes (x, y, z) with an additional timestamp value.
Zeng et al. ~\cite{zeng2014convolutional} interpret each axis of a sensor as a filter of the input to a 1D convolutional layer.
We keep the same procedure but additionally stack each of the four sensors (two accelerometers, two gyroscopes) vertically to create an image-like representation.
Fig.~\ref{fig2} visualizes the shape of our input data.
Height and width of the image-like representation are represented by four sensors and $n$ datapoints.
RGB-like channels are represented by the three axes x, y, and z.
The square in the top left corner visualizes the kernel sliding across the input data.
We split each recording into smaller samples utilizing a sliding window with an overlap of 75\%.
This allows us to create many examples from few data recordings.
In our experiments we test five different window sizes, namely 1000ms, 2000ms, 5000ms, 10000ms, and 20000ms resulting in 25, 50, 125, 250, and 500 data points per example.
This leads to 5937, 2971, 1150, 575, and 286 samples respectively.
For each experiment, we use a 80/20 test/train split in order to evaluate the network's performance on unseen data.

\section{Classification through a 2D convolutional neural network}

\begin{figure}
\centering 
\includegraphics[width=0.70\textwidth]{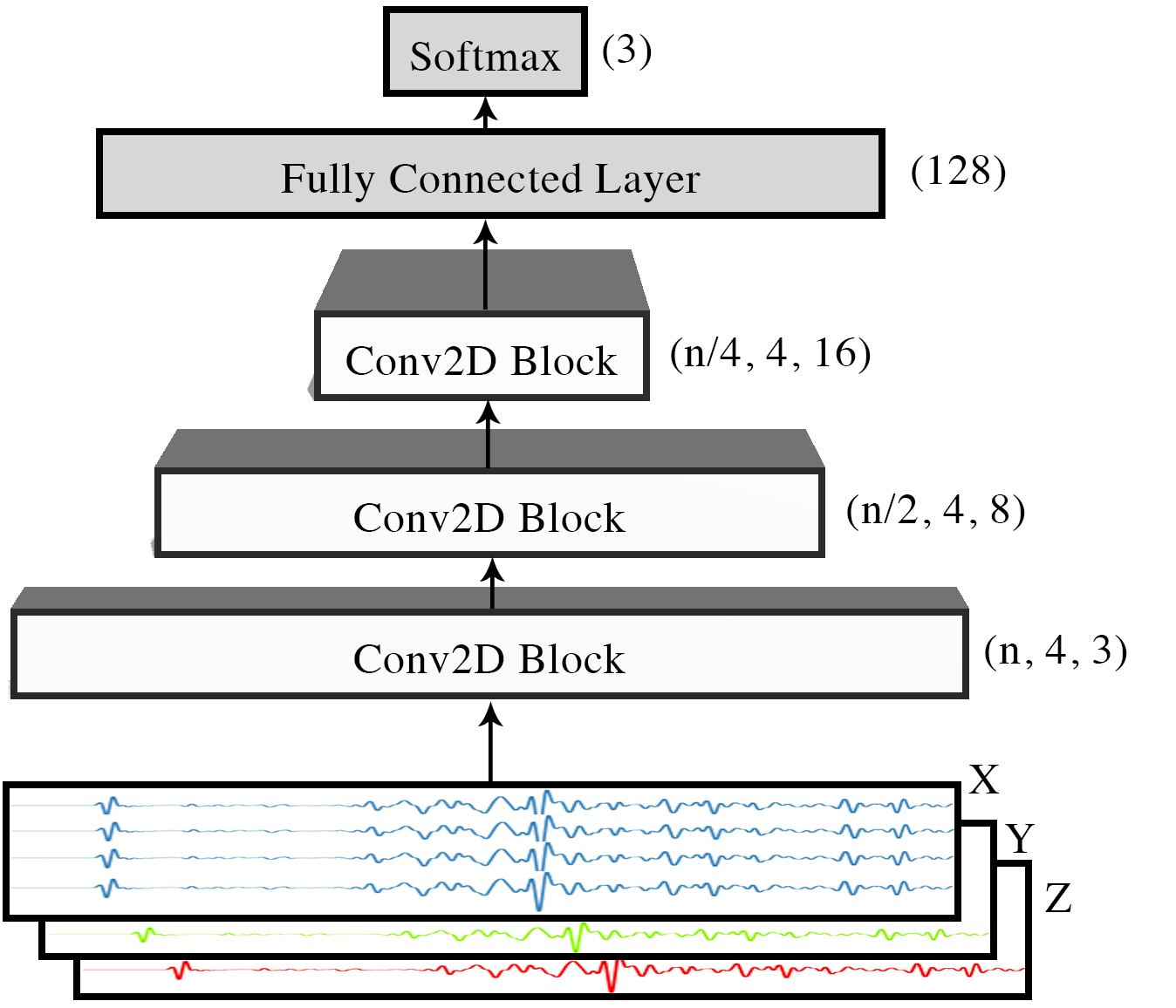}
\caption{Convolutional Neural Network for trail difficulty classification on stacked accelerometer and gyroscope data. 
The input data is of shape ($n$, 4, 3), with $n$ being the amount of data points per sample. 
After the input layer, three convolutional blocks and two fully connected layers follow.}
\label{fig3}	
\end{figure}

In order to classify mountainbike downhill trails regarding their difficulty, we apply a convolutional neural network.
Fig.~\ref{fig3} visualizes the network's architecture.
The input to the first block is of shape ($n$, 4, 3), with $n$ being the amount of data points per sample. 
One sample consists of data of four sensors (vertically stacked), with each three axes (filters), and a sample size of $n$ data points.
We chain three convolutional blocks followed by two Dense Layers.
Each convolutional block consists of one Conv2D ~\cite{krizhevsky2012imagenet}, a Batch Normalization ~\cite{ioffe2015batch}, a ReLU Activation ~\cite{hahnloser2000digital}, a Max Pool ~\cite{matsugu2003subject} and a Dropout Layer ~\cite{srivastava2014dropout}.
The convolutional layers use a kernel of shape ($m$, 2) and a stride of (1, 1), with $m$ being the length of the kernel.
Multiple values for $n$ and $m$ are tested in the experiments.
All convolutional layers use the padding 'same' ~\cite{keraspaddingsame}.
With this setting, the width and height dimensions of the in- and output of a convolutional layer stay the same.
Furthermore, we add L2 regularization to each convolutional layer ~\cite{ng2004feature}.
L2 regularization shifts outlier weights closer to 0.
Max Pool layers use a pool size of (2, 1), which reduces the shape by approximately half in length.
The dropout rate of each Dropout Layer is 0.3.
The Conv2D Layers of the second and third convolutional block have 8 and 16 filters respectively.
After the convolutional blocks, we add two Dense Layers. 
The first layer has 128 units and a ReLU Activation.
The second and final layer has three Softmax activated units, which represent the predicted label.
The network uses the Adam optimizer ~\cite{kingma2014adam} with a learning rate of $0.001$ and a Sparse Categorical Crossentropy as it's loss function.
This configuration proofed to be the best in our experiments.

\subsection{Experiments}

\begin{table}[]
\centering
\setlength{\tabcolsep}{0.2em} 
{\renewcommand{\arraystretch}{1.07}
\begin{tabular}{cc|c|c|c|c|c|c|c|}
\cline{3-9}
                                                                                               &         & \multicolumn{5}{c|}{kernel size}                                                                                                                                                                                                                                                                    & Samples & \begin{tabular}[c]{@{}c@{}}Samples after \\ over-sampling\end{tabular} \\ \cline{3-9} 
                                                                                               &         & (5,2)                                                  & (10,2)                                                 & (20,2)                                                 & (40,2)                                                 & (60,2)                                                          &         &                                                                        \\ \hline
\multicolumn{1}{|c|}{\multirow{5}{*}{\begin{tabular}[c]{@{}c@{}}window\\ size\end{tabular}}} & 1000ms  & \begin{tabular}[c]{@{}c@{}}0.4990\\ (271)\end{tabular} & \begin{tabular}[c]{@{}c@{}}0.5313\\ (124)\end{tabular} & \begin{tabular}[c]{@{}c@{}}0.5165\\ (249)\end{tabular} & -                                                      & -                                                               & 5937    & 10368                                                                  \\ \cline{2-9} 
\multicolumn{1}{|c|}{}                                                                         & 2000ms  & \begin{tabular}[c]{@{}c@{}}0.5155\\ (259)\end{tabular} & \begin{tabular}[c]{@{}c@{}}0.5585\\ (300)\end{tabular} & \begin{tabular}[c]{@{}c@{}}0.5734\\ (247)\end{tabular} & \begin{tabular}[c]{@{}c@{}}0.5599\\ (183)\end{tabular} & -                                                               & 2971    & 5073                                                                   \\ \cline{2-9} 
\multicolumn{1}{|c|}{}                                                                         & 5000ms  & \begin{tabular}[c]{@{}c@{}}0.6181\\ (300)\end{tabular} & \begin{tabular}[c]{@{}c@{}}0.6632\\ (515)\end{tabular} & \begin{tabular}[c]{@{}c@{}}0.7743\\ (726)\end{tabular} & \begin{tabular}[c]{@{}c@{}}0.6632\\ (63)\end{tabular}  & \begin{tabular}[c]{@{}c@{}}0.7778\\ (79)\end{tabular}           & 1150    & 2019                                                                   \\ \cline{2-9} 
\multicolumn{1}{|c|}{}                                                                         & 10000ms & \begin{tabular}[c]{@{}c@{}}0.7222\\ (425)\end{tabular} & \begin{tabular}[c]{@{}c@{}}0.6875\\ (473)\end{tabular} & \begin{tabular}[c]{@{}c@{}}0.7639\\ (548)\end{tabular} & \begin{tabular}[c]{@{}c@{}}0.8681\\ (607)\end{tabular} & \textbf{\begin{tabular}[c]{@{}c@{}}0.9097\\ (781)\end{tabular}} & 575     & 978                                                                    \\ \cline{2-9} 
\multicolumn{1}{|c|}{}                                                                         & 20000ms & \begin{tabular}[c]{@{}c@{}}0.6250\\ (456)\end{tabular} & \begin{tabular}[c]{@{}c@{}}0.6111\\ (5)\end{tabular}   & \begin{tabular}[c]{@{}c@{}}0.6111\\ (162)\end{tabular} & \begin{tabular}[c]{@{}c@{}}0.6250\\ (8)\end{tabular}   & \begin{tabular}[c]{@{}c@{}}0.6806\\ (545)\end{tabular}          & 286     & 498                                                                    \\ \hline
\end{tabular}
}
\caption{Sparse categorical accuracy and amount of epochs of the experiments. A window size of 10000ms with a kernel size of (60,2) creates the highest accuracy (0.9097) after 781 epochs on the test dataset.}
\label{table:configurations}
\end{table}

Due to the fact that there are no established neural network configurations for trail difficulty classification, we evaluate 25 combinations of window- and kernel sizes.
We test five window sizes (1000ms, 2000ms, 5000ms, 10000ms, 20000ms) and five kernel sizes ((5,2), (10,2), (20, 2), (40,2), (60,2)) (see Table~\ref{table:configurations}).
For empty result cells, the amount of datapoints per sample is smaller than the kernel length.
The dataset includes approximately 32\% of samples of label 0, 56\% of samples of label 1 and 12\% of samples of label 2.
This uneven distribution led the model to rarely predict the labels 0 and 2.
Therefore we decided to compensate the inequality by copying existing examples of the underrepresented classes within the training set (so that the classes are balanced equally).
In order to reduce overfitting, we add an early stopping callback to the network, which stops the training process when there is no improvement for 250 epochs (the patience value).
With smaller patience values the network stopped learning too early in some cases.
The maximum amount of epochs for training is 1500.
We run a batch size of 32 and a steady learning rate of $0.001$.
In Table~\ref{table:configurations} we show the resulting Sparse Categorical Accuracy, the amount of epochs before training was stopped and the amount of samples in the train set.
The Sparse Categorical Accuracy measures the accuracy of the result of sparse multiclass classification problems ~\cite{moolayil2019keras}.
For every experiment, we use a sliding window with an overlap of 75\%.
To not have many highly similar examples in one batch, we shuffle the data before training.
Short window sizes (1000ms, 2000ms) show lower accuracy than the larger samples across all kernel sizes.
This could be attributed to the low amount of datapoints within a sample (25) as well as the short sample not representing the subsection of the trail.

\begin{figure}
\vspace{-5mm}
\centering
\includegraphics[width=\textwidth]{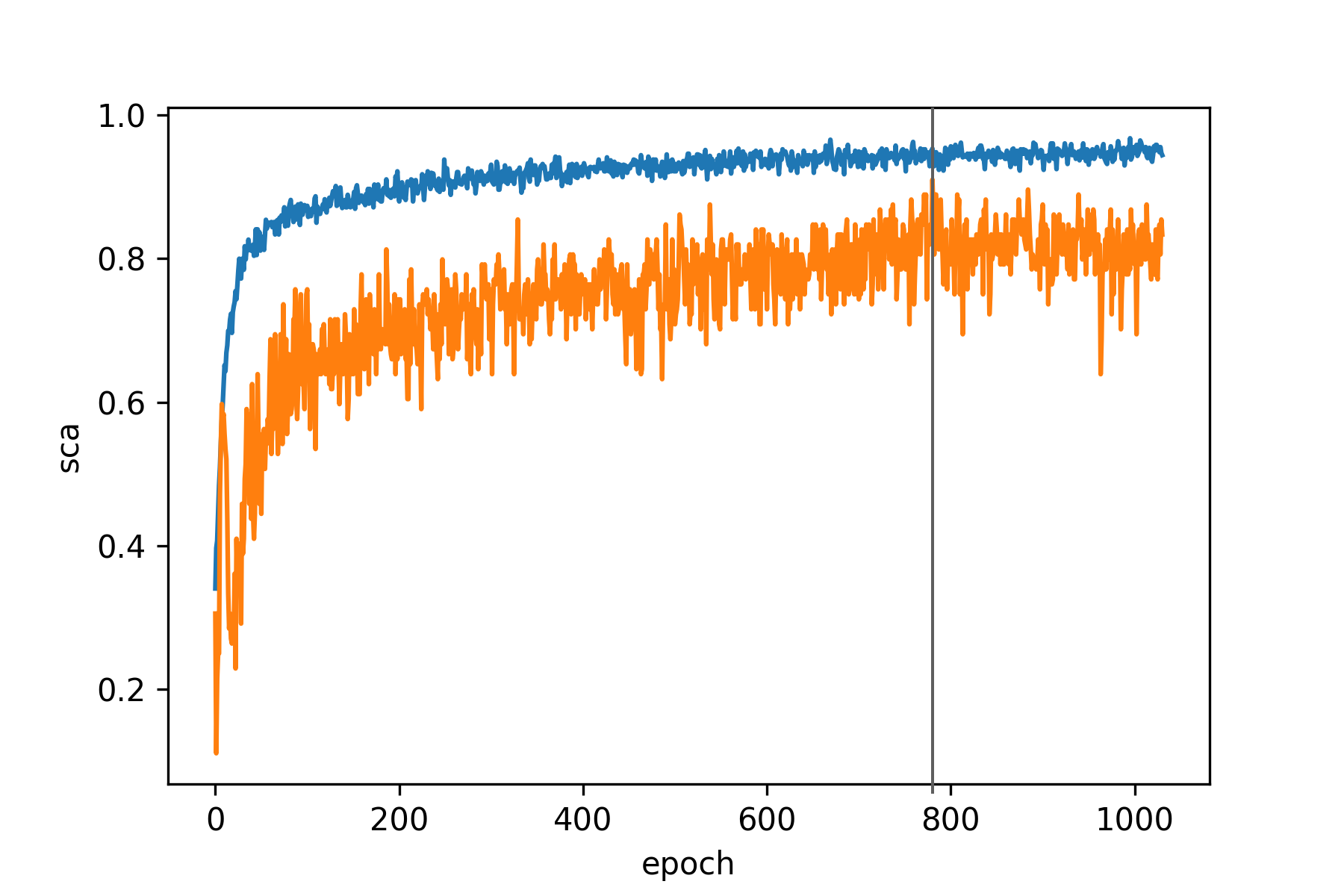}
\caption{Plot of the Sparse Categorical Accuracy (sca) on the train set (blue, upper) as well as on the test set (orange, lower).}
\label{fig4}	
\end{figure} 

\FloatBarrier

The lowest accuracy (0.4990) was reached with window size 1000ms and kernel size (5,2).
With a window size of 10000ms and a kernel size of (60,2), we achieve a high sparse categorical accuracy of 0.9097.
This leads to the conclusion, that a window length of 10000ms is necessary to represent a downhill trail subsection appropriately.
Longer sequential dependencies (by using larger kernel lengths) show a positive effect on the difficulty classification as well.

Fig.~\ref{fig4} shows the curves of the Sparse Categorical Accuracy on the train as well on the test dataset across 1000 epochs.
Both values increase early on and level out with no major overfitting visible in the plot.
The highest accuracy on the test dataset was achieved after 781 epochs.

\begin{figure}
\centering 
\vspace{-5mm}
\includegraphics[width=0.8\textwidth]{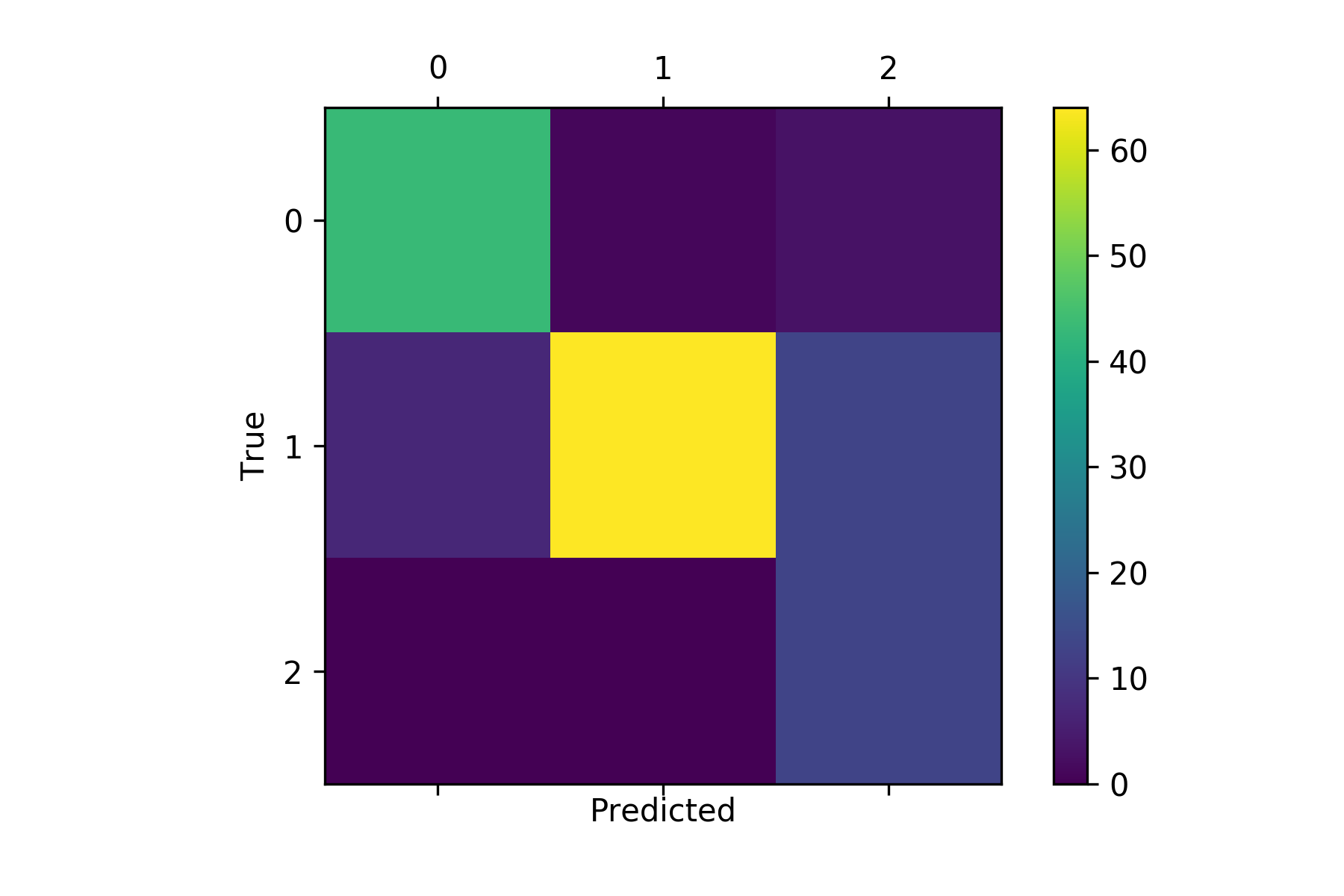}
\caption{The confusion matrix for a window size of 10000ms and a kernel size of (60,2).}
\label{fig5}	
\end{figure}

Fig.~\ref{fig5} shows the confusion matrix of the best resulting configuration, namely a window size of 10000ms and a kernel size of (60,2).
Good results for all three classes are shown, with only few false positives in neighbored areas. 
The matrix also highlights the fact, that the label 2 (hard) is underrepresented.
However, the distribution of correctly predicted labels matches the distribution of the raw dataset well.

\section{Conclusion}
In this work, we proposed an end-to-end deep learning approach to classify mountainbike downhill trails regarding their difficulty. 
We gave an introduction to multiple official difficulty scales and decided to use the \textit{Singletrail-Skala} for this work.
Using mbientlab Meta Motion r0.2 sensor units, we recorded multiple rides on multiple trail segments, resulting in 2971 training examples for the best window-size/kernel-size combination.
The sensor units provided us with accelerometer and gyroscope data in each three axes, which we concatenated to create an image-like representation of the data.
Downhill trails were labeled according to their \textit{Singletrail-Skala} rating and a subjective up- or downgrading for subsections, that strongly diverge from their rating.
We implemented a 2D convolutional neural network with two dense layers at the end for the classification process.
We ran experiments with five different window sizes (1000ms, 2000ms, 5000ms, 10000ms, 20000ms) and five different kernel sizes ((5,2), (10,2), (20,2), (40,2), (60,2)).
The best result could be observed with a sample size of 10000ms and a kernel size of (60,2), resulting in a Sparse Categorical Accuracy of 0.9097 on a 80/20 train/test split. 
In future work, one could think of a non-supervised clustering method to avoid subjective input.
Additionally, the dataset could possibly be improved by using more sensors, like high-resolution barometers or heartrate sensors.
As can be seen in Fig.~\ref{fig5} more examples for hard sections (label 2) are needed.
This category is underrepresented in the data we collected.
\\
With this work, we hope to reduce the amount of subjective rating of mountainbike trails and make their difficulty measurable.
An automated recognition of downhill trail difficulty could be advantageous in diverse scenarios.
For unlabeled, or mislabeled trails, our sensor analysis architecture can generate a fitting label.
This can help tourist areas or mountainbike park operators describe the difficulty of new or existing trails consistently across areas, topographies, or countries.
For social fitness networks like e.g. Strava ~\cite{strava} one could think of an automated difficulty grading of rides (or subsections of rides).
This would extend the existing performance comparison factors, like speed or distance, by a value for downhill trail difficulty.
Furthermore, we hope to promote data analytics in the sport of mountainbiking by releasing a bigger and improved version of our dataset soon.

\bibliographystyle{IEEEtran}
\bibliography{library.bib}

\end{document}